\documentclass[5p,times,twocolumn]{elsarticle}

\usepackage{hyperref}
\usepackage{times}
\usepackage{comment}
\usepackage{epsfig}
\usepackage{graphicx}
\usepackage{subfigure}
\usepackage{amsmath}
\usepackage{amssymb}
\usepackage{multirow}
\usepackage{caption}
\usepackage{colortbl}
\usepackage{algorithm}
\usepackage{algorithmic}  
\usepackage{bm}
\journal{Pattern Recognition}

\newcommand{\re}[1]{{\color{black}{#1}}}
\def\x{x}
\def\y{y}

\def\ie{\emph{i.e.}}
\def\Iset{\mathcal{S_{I}}}
\def\Aset{\mathcal{S_{A}}}
\def\Sset{\mathcal{S}}
\def\w{\bm{w}}
\def\thetabf{{\boldsymbol{\theta}}}
\begin{document}

\begin{frontmatter}

\title{Improving Person Re-identification by Attribute and Identity Learning}
\author[ad1]{Yutian Lin}
\author[ad2]{Liang Zheng}
\author[ad1]{Zhedong Zheng}
\author[ad1]{Yu Wu}
\author[ad1]{Zhilan Hu}
\author[ad3]{Chenggang Yan}
\author[ad1]{Yi Yang\corref{mycorrespondingauthor}}
\address[ad1]{Centre for Artificial Intelligence, University of Technology Sydney}
\address[ad2]{Australian National University}
\address[ad3]{Hangzhou Dianzi University}

\cortext[mycorrespondingauthor]{Yi Yang}
\ead{yi.yang@uts.edu.au}

\begin{abstract}
Person re-identification (re-ID) and attribute recognition share a common target at learning pedestrian descriptions. Their difference consists in the granularity. 
Most existing re-ID methods only take identity labels of pedestrians into consideration. However, we find the attributes, containing detailed local descriptions, are beneficial in allowing the re-ID model to learn more discriminative feature representations.
In this paper, based on the complementarity of attribute labels and ID labels, we propose an attribute-person recognition (APR) network, a multi-task network which learns a re-ID embedding and at the same time predicts pedestrian attributes.
We manually annotate attribute labels for two large-scale re-ID datasets, and systematically investigate how person re-ID and attribute recognition benefit from each other. 
In addition, we re-weight the attribute predictions considering the dependencies and correlations among the attributes.
The experimental results on two large-scale re-ID benchmarks demonstrate that by learning a more discriminative representation, APR achieves competitive re-ID performance compared with the state-of-the-art methods. 
We use APR to speed up the retrieval process by ten times with a minor accuracy drop of 2.92\% on Market-1501.
Besides, we also apply APR on the attribute recognition task and demonstrate improvement over the baselines. 

\end{abstract}


\end{frontmatter}

\section{Introduction}
\begin{figure}[t]
        \begin{center}
            \includegraphics[width=0.9\linewidth]{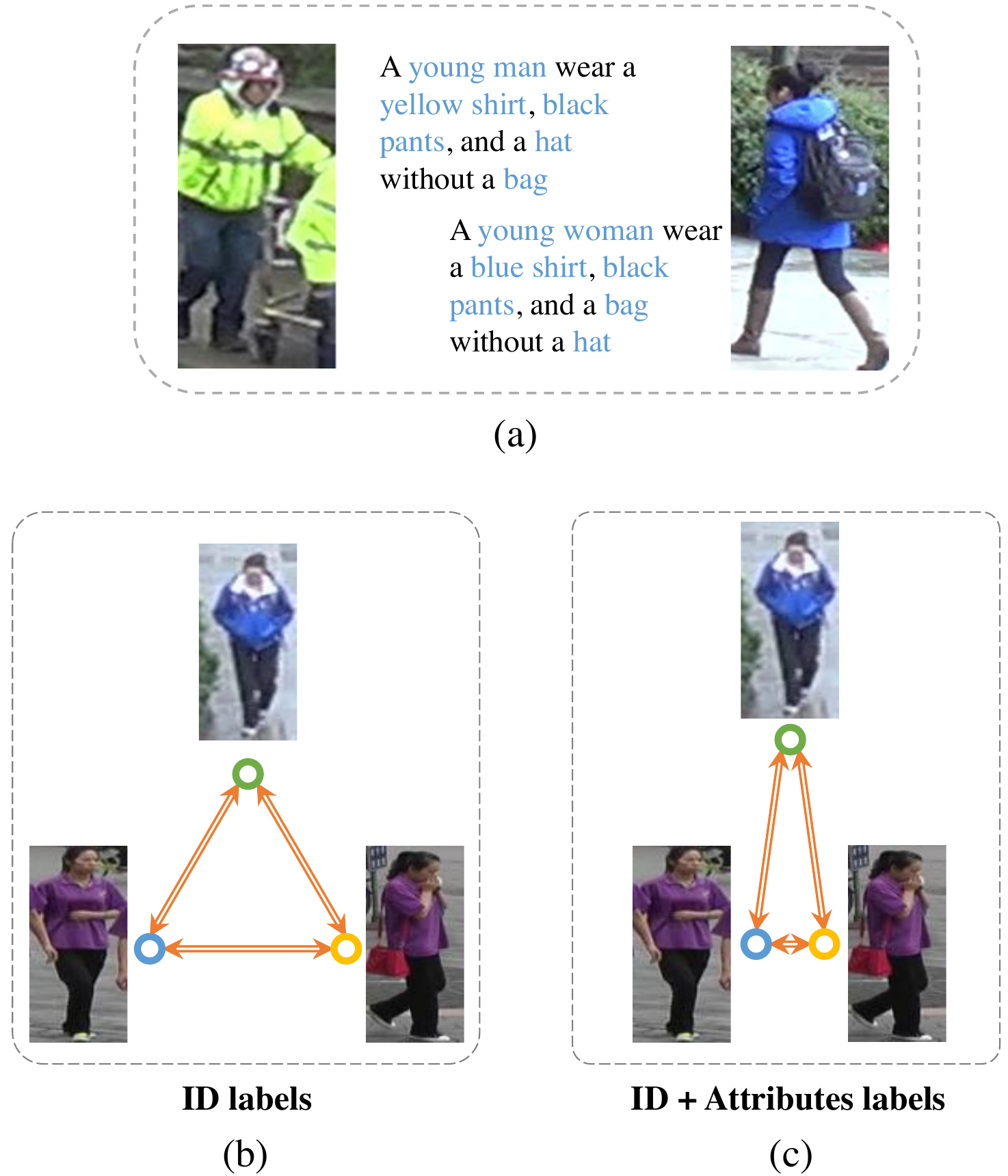}
        \end{center}
        \caption{(a) Two examples of how attributes describe a person. (b) The three pedestrians are of different identities. Guided with only ID labels, images of three different identities have the same label distance between each other. (c) The three pedestrians are of different identities. Guided with ID and attribute labels, the bottom two identities are getting closer to each other in target space while the top one is pushed far away.
        }
        \label{fig:1motivation}
    \end{figure}

Person re-ID \citep{zhu2018fast,liu2017end,wu2017deep,ren2017multi} and attribute recognition \citep{zhu2017multi, deng2014pedestrian, abdulnabi2015multi} both imply critical applications in surveillance. Person re-ID is a task of finding the queried person from non-overlapping cameras, while the goal of attribute recognition is to predict the presence of a set of attributes from an image. Attributes describe detail information for a person, including gender, accessory, the color of clothes, \emph{etc}. Two examples of how attributes describe a person are shown in Fig. \ref{fig:1motivation} (a). In this paper, we aim to improve the performance of large-scale person re-ID, using complementary cues from attribute labels.
The motivation of this paper is that existing large-scale pedestrian datasets for re-ID contains only annotations of identity labels, we believe that attribute labels are complementary with identity labels in person re-ID. 
    
The effectiveness of attribute labels is three-fold:
First, training with attribute labels improves the discriminative ability of a re-ID model. The ID label can only coarsely define the distances among all the identities. This is not optimal since the appearance similarity of identities is overlooked. 
For example, as shown in Fig. \ref{fig:1motivation}(b), the bottom two pedestrians are very similar to each other, and they look very different from the top one. However, with only identity labels, the three pedestrians are uniformly distributed in the target space, which may harm model training. A more natural way is to treat these pedestrians differently according to their similarity. Attribute labels can depict pedestrian images with more detailed descriptions. These local descriptions push pedestrians with similar appearances closer to each other and those different away from each other (Fig. \ref{fig:1motivation}(c)). 
Second, detailed attribute labels explicitly guide the model to learn the person representation by designated human characteristics. With only identity labels and no detailed descriptions, the re-ID model have to infer the differences of pedestrians by itself, which is hard to learn a good semantic feature representation for persons. With the attribute labels, the model is able to learn to classify the pedestrians by explicitly focusing on some local semantic descriptions, which greatly ease the training of models.     
Third, attributes can be used to accelerate the retrieval process of re-ID. The main idea is to filter out some gallery images that do not have the same attributes as the query. 
    
Several datasets are released for the pedestrian attribute. Li \emph{et.al} \cite{li2016richly} release a large-scale pedestrian attribute dataset RAP. Since the RAP dataset does not have ID labels, it is usually used to transfer attribute knowledge to the target re-ID dataset.
In \cite{deng2014pedestrian}, the PETA dataset is proposed which contains both attribute and identity attributes. However, PETA is comprised of small datasets and most of the datasets only contain one or two images for an identity. The lack of training images per identity limits the deep learning research. When using attributes for re-ID, attributes can be used as auxiliary information for low level features \cite{layne2014re} or used to better match images from two cameras \cite{khamis2014joint,su2017attributes,su2018multitask}. 
 In recent years, some deep learning methods are proposed \cite{franco2017convolutional,su2018multitype,schumann2017person}. In these works, the network is usually trained by several stages. Franco \emph{et al.} \cite{franco2017convolutional} propose a coarse-to-fine learning framework. The network is comprised of a set of hybrid deep networks, and one of the networks is trained to classify the gender of a person. In this work, the networks are trained separately and thus may overlook the complementarity of the general ID information and the attribute details. Besides, since gender is the only attribute used in the work, the correlation between attributes is not leveraged in \cite{franco2017convolutional}. In \cite{su2018multitype,schumann2017person}, the network is first trained on an independent attribute dataset, and then the learned information is transferred to the re-ID task.
 A work closest to ours consists of \cite{matsukawa2016person}, in which the CNN embedding is only optimized by the attribute loss. We will show that by combining the identification and attribute recognition with an attribute re-weighting module, the APR network is superior to the method proposed in \cite{matsukawa2016person}. 

Comparing with previous methods, our paper differs in two main aspects. First, our work systematically investigates how person re-ID and attribute recognition benefit each other by a jointly learned network. On the one hand, identity labels provide global descriptions for person images, which have been proved effective for learning a good person representation in many re-ID works \cite{Zheng_2015_ICCV,chen2016deep,wu2018cvpr_oneshot}. On the other hand, attribute labels provide detailed local descriptions. By exploiting both local (attribute) and global (identity) information, one is able to learn a better representation for a person, thereby achieving higher accuracy for person attribute recognition and person re-ID. Second, in previous works, the correlations of attributes are hardly considered. In fact, many attributes usually co-occur for a person, and the correlations of attributes may be helpful to re-weight the prediction of each attribute. For example, the attributes ``skirt” and ``handbag” are highly related to ``female” rather than ``male”. Given these gender-biased attribute descriptions, the probability of the attribute ``female” should increase. We thereby introduce an Attribute Re-weighting Module to utilize correlations among attributes and optimize attribute predictions.

In this paper, we propose the attribute-person recognition (APR) network to exploit both identity labels and attribute annotations for person re-ID.
By combining the attribute recognition task and identity classification task, the APR network is capable of learning more discriminative feature representations for pedestrians, including global and local descriptions.
Specifically, we take attribute predictions as additional cues for the identity classification.
Considering the dependencies among pedestrian attributes, we first re-weight the attribute predictions and then build identification upon these re-weighted attributes descriptions. 
The attribute is also used to speed up the retrieval process by filtering out the gallery images with different attribute from the query image. In the experiment, we show that by applying the attribute acceleration process, the evaluation time is saved to a significant extent.
We evaluate the performance of the proposed method APR on two large-scale re-ID datasets and an attribute recognition dataset. The experimental results show that our method achieves competitive re-ID accuracy to the state-of-the-art methods. 
In addition, we demonstrate that the proposed APR yields improvement in the attribute recognition task over the baseline in all the testing datasets.

Comparing with existing works, our contributions are summarized as follows:
    
(1) We have manually labeled a set of pedestrian attributes for the Market-1501 dataset and the DukeMTMC-reID dataset. Attribute annotations of both datasets are publicly available on our website (\url{https://vana77.github.io}).
        
(2) We propose a novel attribute-person recognition (APR) framework. It learns a discriminative Convolutional Neural Network (CNN) embedding for both person re-identification and attributes recognition. 
    
(3) We introduce the Attribute Re-weighting Module (ARM), which corrects predictions of attributes based on the learned dependency and correlation among attributes.

(4) We propose an attribute acceleration process to speed up the retrieval process by filtering out the gallery images with different attribute from the query image. The experiment shows that the size of the gallery is reduced by ten times, with only a slight accuracy drop of 2.92\%.
    
(5) We achieve competitive accuracy compared with the state-of-the-art re-ID methods on two large-scale datasets, \emph{i.e.}, Market-1501 \citep{Zheng_2015_ICCV} and DukeMTMC\_reID \citep{zheng2017discriminatively}. We also demonstrate improvements in the attribute recognition task. 
    
\section{Related Work}
\textbf{CNN-based person re-ID.}
CNN-based methods are dominating the re-ID community upon the success of deep learning \cite{he2016deep,fan18unsupervisedreid,varior2016gated,zheng2017discriminatively,Xiao_2016_CVPR, zhou2018deep,zhu2017uncovering}.
A branch of works learning deep metrics \cite{ma2014person,li2014deepreid,ding2015deep} that image pairs or triplets are fed into the network. Usually, the spatial constraints are integrated into the similarity learning process \cite{li2014deepreid,ahmed2015improved}. For example, in \cite{varior2016gated}, a gating function is inserted in each convolutional layer, so that some subtle difference between two input images can be captured. 
Generally speaking, deep metric learning methods have advantages in training on relatively small datasets, but its efficiency on larger galleries may be compromised.
Another branch of works learning deep representations \cite{zheng2017discriminatively,Xiao_2016_CVPR, zhou2018deep, wu2018deep}.
Xiao \emph{et al.} \cite{Xiao_2016_CVPR} propose to learn a generic feature embedding by training a classification model from multiple domains with a domain guided dropout.  
In \cite{zheng2017discriminatively}, the combination of verification and classification losses is proven effective. 
Xu \emph{et al.} \cite{xu2018attention} propose a Pose guided Part Attention (PPA)is learned to extract attention-aware feature for body parts from a base network. Then the features of body parts are further re-weighted, resulting in the final feature vector.
Since GAN proposed by Goodfellow \emph{et al.} \cite{goodfellow2014generative}, methods utilizing GAN \cite{Wei_2018_CVPR,Zheng_2017_ICCV} have been proposed to tackle re-ID. In \cite{Wei_2018_CVPR}, a Person Transfer Generative Adversarial Network (PTGAN) is proposed to transfer the image style from one dataset to another while keeping the identity information to bridge the domain gap.   
In \cite{martinel2017person}, a dictionary-learning scheme is applied to transfer the feature learned by object recognition and person detection (source domains) to person re-ID (target domain).
Recently, some semi-supervised methods~\cite{wu2019progressive,wu2018cvpr_oneshot} and unsupervised methods~\cite{lin2019aBottom,fan18unsupervisedreid} has been proposed to address the data problem for re-ID. These methods achieve surprising performances with less or none of annotations. Attributes information also benefits these methods in the semi-supervised task.
    
In this paper, we adopt the simple classification model as our baseline and further exploit the mutual benefit between the traditional identity label and the attribute label.

\textbf{Attributes for person re-ID.}
In some early attempts, attributes are used as auxiliary information to improve low-level features \cite{su2017attributes,su2018multitask,layne2012person,liu2012attribute}. 
In \cite{layne2014re,layne2012person}, low-level descriptors and SVM are used to train attribute detectors, and the attributes are integrated by several metric learning methods. 
Su \emph{et al.} \cite{su2017attributes,su2018multitask}  utilize both low-level features and camera correlations learned from attributes for re-identification in a systematic manner.
In \cite{peng2016joint}, a dictionary learning model is proposed that exploits the discriminative attributes for the classification task. 
Recently, some deep learning methods are proposed.
Franco \emph{et al.} \cite{franco2017convolutional} propose a coarse-to-fine learning framework, which is comprised of a set of hybrid deep networks. The network is trained for distinguishing person/not person, predicting the gender of a person and person re-ID, respectively. In this work, the networks are trained separately and might overlook the complementarity of the ID label and the attribute label. Besides, gender is the only attribute used in the work, so that the correlation between attributes is not leveraged.
However, these works do not consider the correlation between attributes nor show if the proposed method improves the attribute recognition baselines.
In \cite{su2018multitype}, Su \emph{et al.} first train a network on an independent dataset with attribute label, and then fine-tune the network the target dataset using only identity label with triplet loss. Finally, the predicts attribute labels for the target dataset is combined with the independent dataset for the final round of fine-tuning. 
Similarly, in \cite{schumann2017person}, the network is pre-trained on an independent dataset labeled with attributes, and then fine-tuned on another set with person ID.
In \cite{ijcai2018-153}, a set of attribute labels are used as the query to retrieve the person image. Adversarial learning is used to generate image-analogous concepts for query attributes and get it matched with the image in both the global level and semantic ID level.
The attribute is also used as supervision for unsupervised learning. 
Wang \emph{et al.} \cite{Wang_2018_CVPR} propose an unsupervised re-ID method that shares the source domain knowledge through attributes learned from labelled source data and transfers such knowledge to unlabelled target data by a joint attribute identity transfer learning across domains.

\begin{figure}[t]
    \begin{center}
    \includegraphics[width=\linewidth]{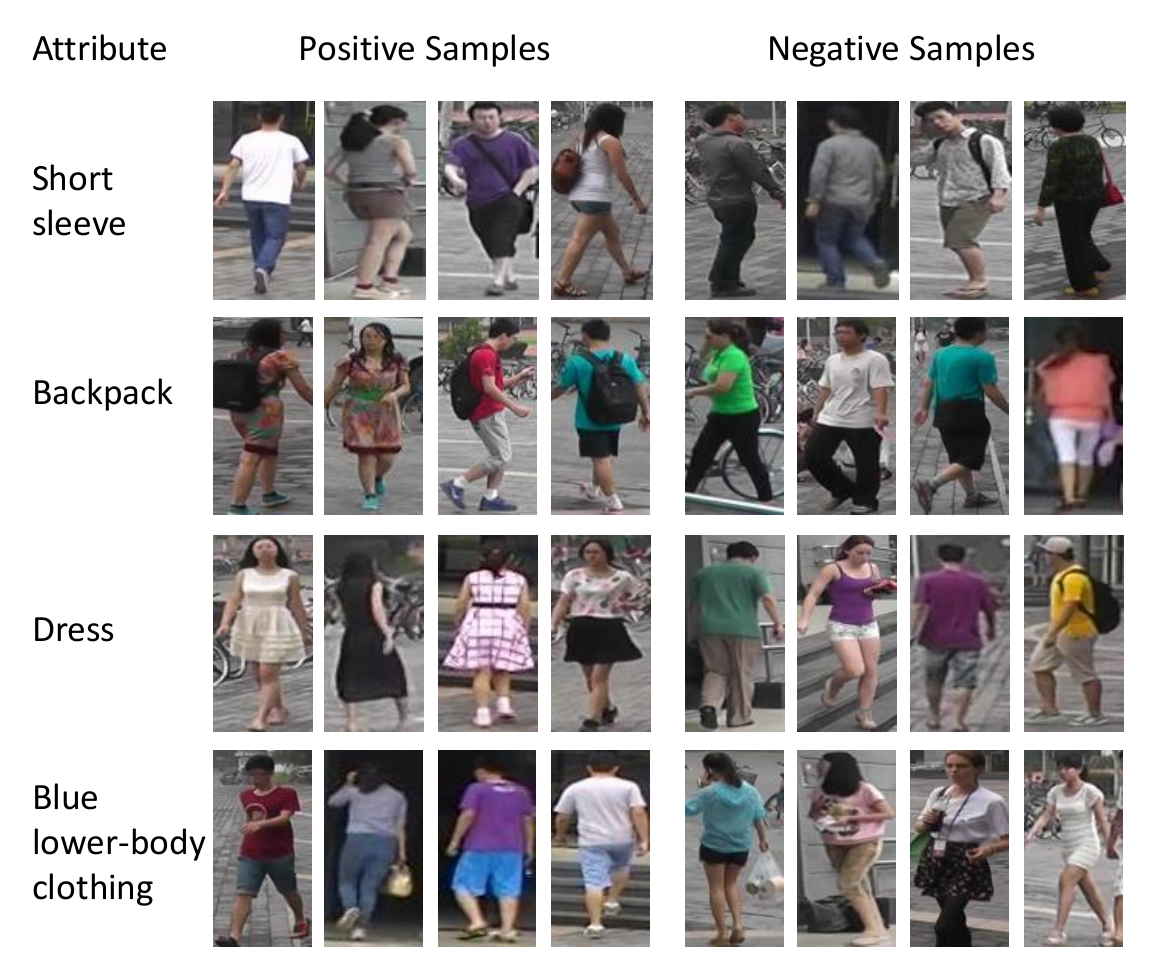}
    \end{center}
    \caption{Positive and negative examples of some representative attributes: \emph{short sleeve}, \emph{backpack}, \emph{dress}, \emph{blue lower-body clothing}.}
    \label{fig:4attribute_train}
\end{figure}
    
\begin{figure*}[t]
    \begin{center}
    \includegraphics[width=0.8\linewidth]{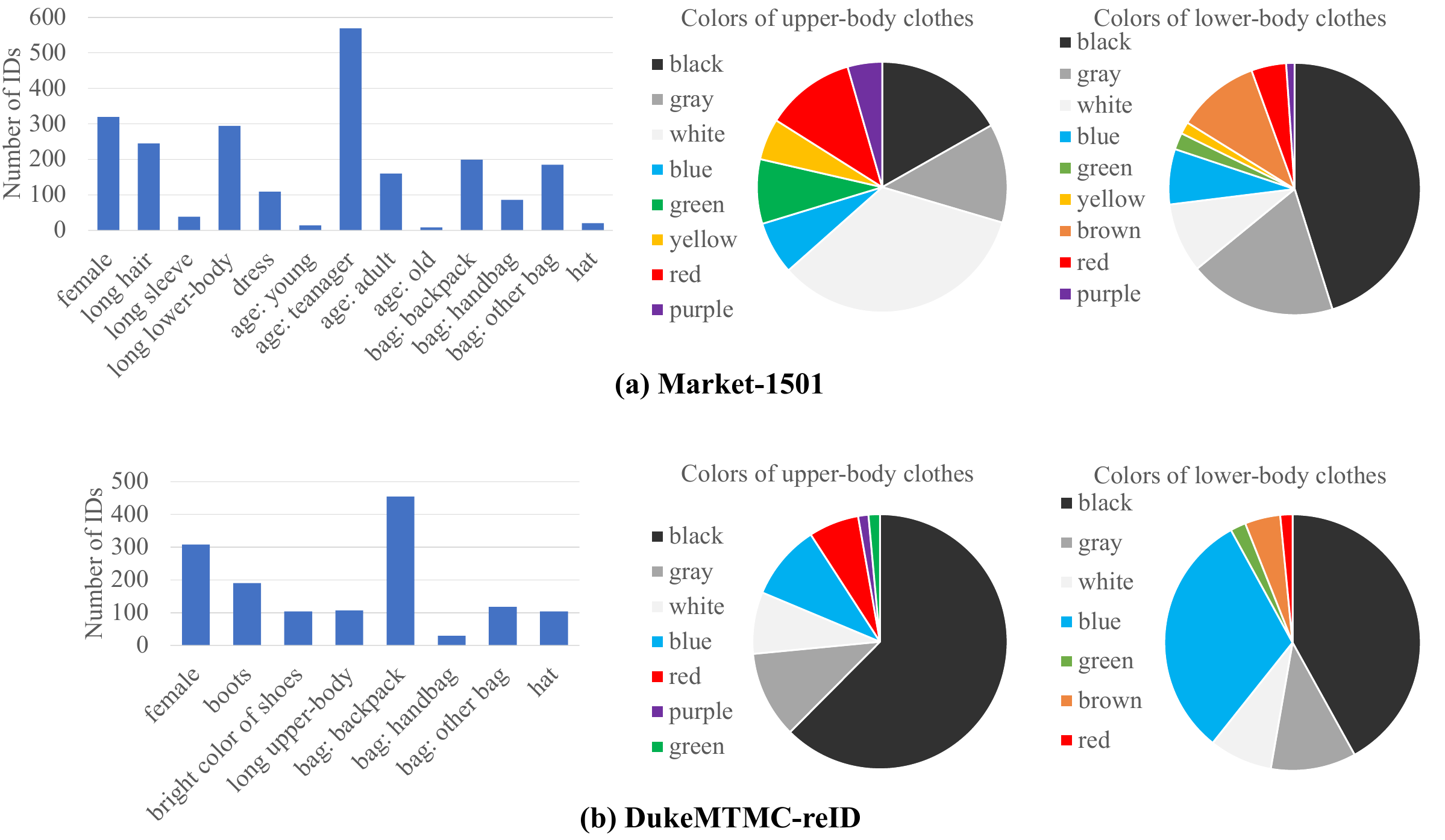}
    \end{center}
    \vspace{-5mm}
    \caption{The distributions of attributes on (a) Market-1501 and (b) DukeMTMC-reID. The left figure of each row shows the numbers of positive IDs for attributes except the color of upper/lower-body clothing. The middle and right pie chart illustrate the distribution of the colors of upper-body clothing and lower-body clothing, respectively. }\label{fig:attribute_distribution}
    \end{figure*}

\begin{figure*}[!t]
    \begin{center}
    \includegraphics[width=0.8\linewidth]{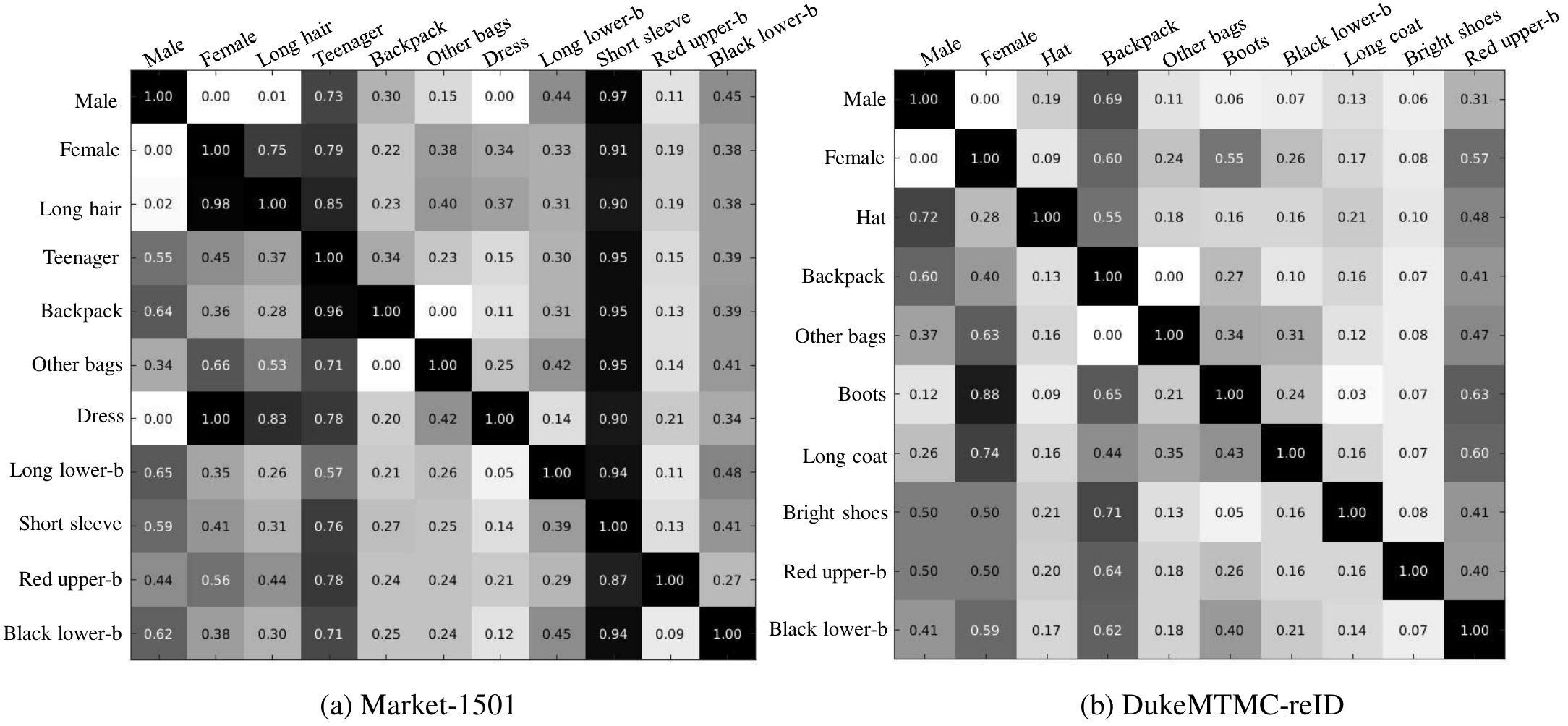}
    \end{center}
    \vspace{-5mm}
    \caption{Attribute correlations on the Market-1501 and DukeMTMC-reID datasets. A larger value indicates a higher correlation between the two attributes. We only show some of the representative attributes in the figure. }\label{fig:attribute_matrix}
\end{figure*} 

\section{Attribute Annotation}
We manually annotate the Market-1501 \cite{Zheng_2015_ICCV} dataset and the DukeMTMC-reID \cite{Zheng_2017_ICCV} dataset with attribute labels. Although the Market-1501 and DukeMTMC-reID datasets are both collected on university campuses, and most identities are students, they are significantly different in seasons (summer vs. winter) and thus have distinct clothes. For instance, many people wear dresses or shorts in Market-1501, but most of the people wear pants in DukeMTMC-reID. So for the two datasets, we use two different sets of attributes. The attributes are carefully selected considering the characteristics of the datasets, so that the label distribution of an attribute (\emph{e.g.}, wearing a hat or not) is not heavily biased. 
    
For Market-1501, we have labeled 27 attributes: gender (male, female), hair length (long, short), sleeve length (long,  short), length of lower-body clothing (long,  short), type of lower-body clothing (pants, dress), wearing hat (yes, no), carrying backpack (yes, no), carrying handbag (yes, no), carrying other types of bag (yes, no), 8 colors of upper-body clothing (black, white, red, purple, yellow, gray, blue, green), 9 colors of lower-body clothing (black, white, red, purple, yellow, gray, blue, green, brown) and age (child, teenager, adult, old). Positive and negative examples of some representative attributes of the Market-1501 dataset are shown in Fig. \ref{fig:4attribute_train}.
    
For DukeMTMC-reID, we have labeled 23 attributes: gender (male, female), shoe type (boots, other shoes), wearing hat (yes, no), carrying backpack (yes, no), carrying handbag (yes, no), carrying other types of bag (yes, no), color of shoes (dark, bright), length of upper-body clothing (long, short), 8 colors of upper-body clothing (black, white, red, purple, gray, blue, green, brown) and 7 colors of lower-body clothing (black, white, red, gray, blue, green, brown). 
    
Note that all the attributes are annotated at the identity level. For example, in Fig. \ref{fig:4attribute_train}, the first two images in the second row are of the same identity. Although we cannot see the backpack clearly in the second image, we still annotate there is a ``backpack'' in the image.
For both Market-1501 and DukeMTMC-reID, we illustrate the attribute distribution in Fig \ref{fig:attribute_distribution}. 
We define correlation of two attributes as the possibility that they co-occur on a person.
We show the correlations between some representative attributes in Fig \ref{fig:attribute_matrix}. Attribute pairs with higher correlation are in a darker grid.

\section{The Proposed Method}
We first describe the necessary notations and two baseline methods in Section \ref{sec:Preliminaries} and then introduce our proposed Attribute-Person Recognition network in Section \ref{sec:APR}. Finally, we introduce the attribute acceleration process in Section \ref{sec:accelerate}.
    
\subsection{Preliminaries}\label{sec:Preliminaries}

Let $\Iset = \{ (\x_{1}, \y_{1}), ..., (\x_{n}, \y_{n})\}$ be the pedestrian identity labeled data set, where $\x_{i}$ and $\y_i$ denotes the $i$-th image and its identity label, respectively.
For each image $\x_{i} \in \Iset$, we have the attributes annotations $\bm{a}_{i} = (a^{1}_{i}, a^{2}_{i}, ..., a^{m}_{i})$, where $a^{j}_{i}$ is the $j$-th attribute label for the image $\x_{i}$, and $m$ is the number of attributes classes. 
Let $\Aset = \{(\x_{1}, \bm{a}_{1}), ..., (\x_{n}, \bm{a}_{n})\}$ be the attribute labeled set.
Note that set $\Iset$ and set $\Aset$ share common pedestrian images $\{\x_i\}$.
Based on these two set $\Iset$ and $\Aset$, we have the following two baselines:
    
\textbf{Baseline~1 ID-discriminative Embedding (IDE).} 
Following \cite{Zheng_2015_ICCV}, we take IDE to train the re-ID model, which regards re-ID training process as an image identity classification task.
It is trained only on the identity label data set $\Iset$. We have the following objective function for IDE:

{
\begin{align}\label{eq:IDE_objective_function}
  \min_{\thetabf_{I}, \w_{I}} \sum_{i=1}^{n} \ell( f_{I}(\w_{I}; \phi(\thetabf_{I}; \x_{i})), \y_{i} )  ,
\end{align}
}

\noindent
where $\phi$ is the embedding function, parameterized by $\thetabf_{I}$, to extract the feature from the data $\x_{i}$. CNN models \cite{Zheng_2015_ICCV,zheng2017discriminatively} are usually used as the embedding function $\phi$.
$f_I$ is an identity classifier, parameterized by $\w_{I}$, to classify the embedded image feature $\phi(\thetabf_{I}; \x_{i})$ into a $k$-dimension identity confidence estimation, in which $k$ is the number of identities. 
$\ell$ denotes the suffered loss between classifier prediction and its ground truth label.

\textbf{Baseline 2 Attribute Recognition Network (ARN).} Similar to the IDE baseline for identity prediction, we propose the Attribute Recognition Network (ARN) for attribute prediction. 
ARN is trained only on the attribute label data set $\Aset$. We define the following objective function for ARN:

{
\begin{align}\label{eq:ARN_objective_function}
\min_{\thetabf, \w_{A}} \sum_{i=1}^{n} \sum_{j=1}^{m} \ell( f_{A_j}(\w_{A_j}; \phi(\thetabf; \x_{i})), a^{j}_{i} ),
\end{align}
}

\noindent
where $f_{A_j}$ is the $j$-th attribute classifier, parameterized by $\w_{A_j}$, to classify the embedded image representation $\phi(\thetabf; \x_{i})$ to the $j$-th attribute prediction. We take the sum of all the suffered losses for $m$ attribute predictions on the input image $x_{i}$ as the loss for the $i$-th sample.

In the evaluation stage of person re-ID task, for both baseline models, we use the embedding function $\phi(\thetabf; \cdot)$ to embed the query and gallery images into the feature space. 
The query result is the ranking list of all gallery data according to the Euclidean Distance between the query data and each gallery data, \ie., ${||\phi(\thetabf; \x_{q}) - \phi(\thetabf; \x_{g})||}_2$, where $\x_{q}$ and $\x_{g}$ denote the query image and the gallery image, respectively.
For the evaluation of attribute recognition task, we take the attribute prediction $f_{A}(\w_{A}; \phi(\thetabf; \cdot))$ as the output, thereby evaluated with the ground truth by the classification metric.

\begin{figure*}[t]
        \begin{center}
            \includegraphics[width=0.85\linewidth]{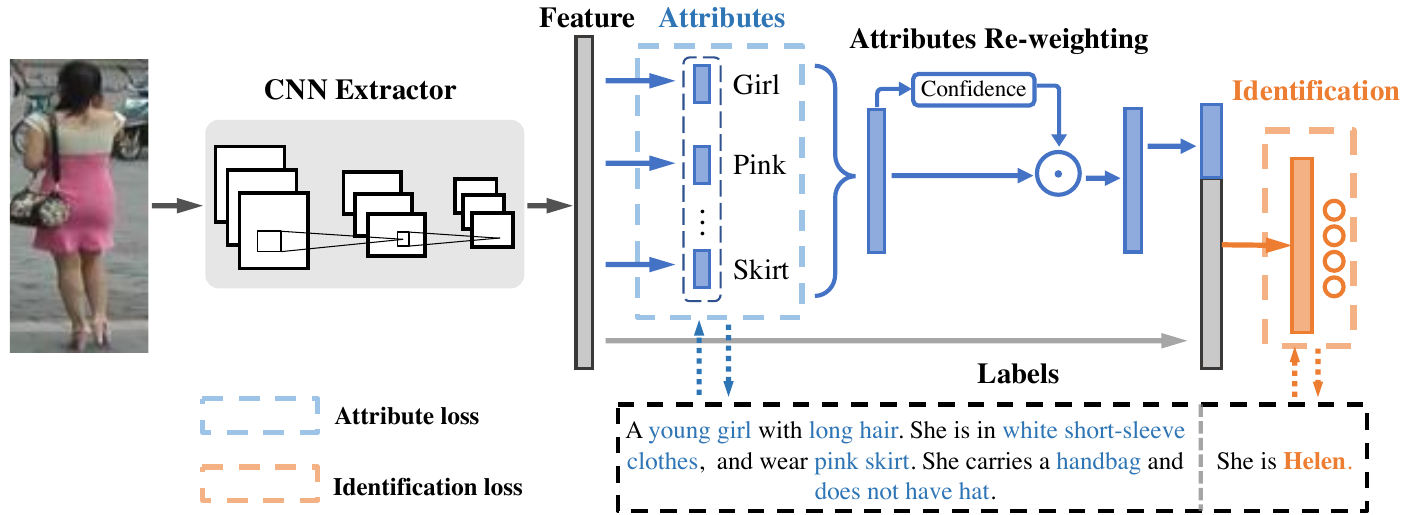}
        \end{center}
\caption{An overview of the APR network. APR contains two classification part, one for attribute recognition and the other for identification. 
Given an input image, the person feature representation is extracted by the CNN extractor $\phi$. Subsequently, the attribute classifiers predict attributes based on the image feature. Here we calculate the attribute classification losses by the attribute predictions and ground truth labels.
For the identity classification part, we take the attribute predictions as additional cues.
Specifically, we first re-weight the local attribute predictions by the Attribute Re-weighting Module and then concatenate them with the global image feature.
The final identification is built upon the concatenated local-global feature.}
        \label{fig:2pipeline}
    \end{figure*}

\subsection{Attribute-Person Recognition Network}\label{sec:APR}

\subsubsection{Architecture Overview} 
The pipeline of the proposed APR network is shown in Fig. \ref{fig:2pipeline}. APR network contains two prediction parts, one for attribute recognition task and the other for the identity classification task. 
Given an input pedestrian image, the APR network first extracts the person feature representation by the CNN extractor $\phi$. Subsequently, APR predicts attributes based on the image feature. Here we calculate the attribute losses by the attribute prediction and ground truth labels. 
For the identity classification part, motivated by the fact that local descriptors (attributes) benefit global identification, we take the attribute predictions as additional cues for identity prediction.
Specifically, to better leverage the attributes, given an input image, the APR network firstly computes attribute losses for the $M$ individual attributes. Then the $M$ prediction scores are concatenated and fed into an Attribute Re-weighting Module (ARM). The output of ARM is then concatenated with the global image feature for ID loss computation. The final identification is built upon the concatenated local-global feature.
    
\subsubsection{Attribute Re-weighting Module} 
Suppose the set of attribute predictions for the image $x$ is $\left\{\tilde{a}^{1}, \tilde{a}^{2}, ..., \tilde{a}^{m}\right\}$, where $\tilde{a}^{j} \in [0,1] $ is the $j$-th attribute prediction score from the attribute classifier $f_{A_j}$. 
We concatenate the prediction scores as vector $\tilde{\bm{a}}$, where $\tilde{\bm{a}} \in \mathbb{R}^{1 \times m}$. Then the confidence score $\bm{c}$ for its prediction $\tilde{\bm{a}}$ is learned as,
{
\begin{align}
  \bm{c} = \texttt{Sigmoid}(\bm{v} \tilde{\bm{a}}^T + \bm{b}),
\end{align}
}
\noindent
where $\bm{v} \in \mathbb{R}^{m \times m}$ and $\bm{b} \in \mathbb{R}^{m \times 1}$ are trainable parameters, and the confidence score $\bm{c} \in \mathbb{R}^{m \times 1}$ is a set of learned weight. 
Therefore, the attribute re-weighting module transforms the original prediction $\tilde{\bm{a}}$ to a new prediction score as
{
\begin{align}\label{eq:re-weight}
  \bm{a} =\bm{c} \circ  \tilde{\bm{a}}^T,
\end{align}
}
\noindent
where $\circ$ is the element-wise multiplication. The re-weighted prediction score $\bm{a}$ is then concatenated with the global image representation for further identity classification.

The motivation behind the Attribute Re-weighting Module (ARM) is to recalibrate the strengths of different activations of the attributes with a general consideration on all attributes.
Therefore, we use trainable parameters ($\bm{v}$, $\bm{b}$) and the Sigmoid activation to perform a gating mechanism on the attribute predictions.
With ARM, the model could learn to utilize the correlation between attributes. For instance, when the prediction scores of ``pink upper-body clothes'' and ``long hair'' are very high, the network may tend to up-weight the prediction scores for the attribute ``female''.

\subsubsection{Optimization}
To exploit the attributes data $\Aset$ as auxiliary annotations for the re-ID task, we propose Attribute-Person Recognition (APR) network. The APR network is trained on the combined data set $\Sset$ of the identity set $\Iset$ and the attribute set $\Aset$, \ie, $\Sset=\{(\x_{1}, \y_{1}, \bm{a}_{1}), ..., (\x_{n}, \y_{n}, \bm{a}_{n})\}$.
For a pedestrian image $\x_{i}$, we first extract the image feature representation by the embedding function $\phi(\thetabf; \cdot)$. Based on the image representation $\phi(\thetabf; \x_i)$, two objective functions are optimized simultaneously:

\textbf{The objective function for attribute predictions.} Similar to the baseline ARN, the attribute predictions are obtained by a set of attribute classifiers on the input image feature, \ie, $\{f_{A_j}(\w_{A_j}; \phi(\thetabf; \x_{i}))\}$. We then optimize the objective function for attribute predictions the same as Eq. \eqref{eq:ARN_objective_function}.
    
\textbf{The objective function for identification.}
To introduce the attributes into identity prediction, we gather the attribute predictions $\{f_{A_j}(\w_{A_j}; \phi(\thetabf; \x_{i}))\}$ and re-weight them by the Attribute Re-weighting Module. We combine the re-weighted attribute predictions $\bm{a}_{i}$ and the image global feature $\phi(\thetabf; \x_{i})$ to form a local-global representation.
The identity classification is built upon the new feature. Thus we have the following objective function for identity prediction:
    
    \begin{equation}\label{eq:APR_ID_objective_function}
\begin{aligned}
\min_{\thetabf, \w_{I}} \sum_{i=1}^{n} \ell( \hat{f}_{I}(\hat{\w}_{I}; \hat{\bm{a}}_{i}, \phi(\thetabf; \x_{i})), \y_{i} )  ,
\end{aligned}
\end{equation}

\noindent
where $\hat{\bm{a}}_{i} = (\hat{a}^{1}_{i}, \hat{a}^{2}_{i}, ..., \hat{a}^{m}_{i})$ is the concatenation of the re-weighted attribute predictions.
$\hat{f}_{I}$ is the identity classier, parameterized by $\hat{\w}_{I}$, to predict the identity based on attribute predictions $\hat{\bm{a}}_{i}$ and image embeddings $\phi(\thetabf; \x_{i})$.

\textbf{The overall objective function.} Considering both attribute recognition and identity prediction, we define the overall objective function as followings:

\begin{equation}\label{eq:APR_objective_function}
\begin{aligned}
\min_{\thetabf, \w_{I}, \w_{A}} &\lambda \sum_{i=1}^{n} \ell( \hat{f}_{I}(\hat{\w}_{I}; \hat{\bm{a}}_{i}, \phi(\thetabf; \x_{i})), \y_{i} )\\
+ &(1-\lambda) \frac{1}{m} \sum_{i=1}^{n} \sum_{j=1}^{m} \ell( f_{A_j}(\w_{A_j}; \phi(\thetabf; \x_{i})), {a}^{j}_{i}  ),
\end{aligned}
\end{equation}

\noindent
where $\lambda$ is a hyper-parameter to balance the identity classification loss and the attribute recognition losses. We empirically discuss the effectiveness of $\lambda$ in Section.\ref{sec:exp_reid}.

\subsection{Attribute acceleration process}\label{sec:accelerate}
In the real-world application, calculating the distance for retrieval has become the main cost for a re-ID system, which is unaffordable. 
Attributes can be used to speed up the evaluation process by filtering the gallery data based on attribute predictions. The main idea is to filter out some gallery images that do not have the same attributes as the query.

During off-line computation, we apply feature extraction and attribute prediction for the gallery images. We take the attribute predictions with high confidence values as reliable ones for both query and gallery images. Then we remove those gallery candidates whose reliable attributes are different from the query. It is clear that the predicted attribute tends to be reliable as the prediction score gets higher. Specifically, we denote $\tau$ to be the threshold value. When the confidence score is higher than $\tau$, the attribute is taken as a reliable one. When an attribute is reliable for both the query and gallery image, we check if the two images have the same prediction on that attribute. If not, this candidate image is removed from the gallery pool.

In real-life applications, this threshold is a trade-off between efficiency and accuracy.
An \textit{aggressive} choice is to set the threshold to a very small value (close to 0). It removes most of the candidates and maintains only a few candidates in on-line matching. This is suitable for the application where the retrieval speed is the main focus. A \textit{conservative} option is to set the threshold to a large value (close to 1). It means we only remove a few candidates that are different in the very reliable attribute predictions from the query. In the empirical studies on Market-1501, we speedup the retrieval process by over ten times with a minor accuracy drop of 2.92\% by setting the threshold to 0.7.

\section{Experimental results}
\subsection{Datasets and Evaluation Protocol}
We conduct experiments on two large-scale person re-ID datasets Market-1501 \cite{Zheng_2015_ICCV} and DukeMTMC-reID \cite{Zheng_2017_ICCV} and one attribute recognition dataset PETA \cite{deng2014pedestrian}. 

\textbf{The Market-1501 dataset} contains 19,732 images for 751 identities for training and 13,328 images for 750 identities for testing. For each image, 27 attributes are annotated. To validate the hyper-parameter $\lambda$ in Eq.\eqref{eq:APR_objective_function}, we use 651 identities in training set for training and the other 100 identities are used as the validation set to determine the value of parameter $\lambda$. We then use this hyper-parameter in the normal 751/750 split. 
    
\textbf{The DukeMTMC-reID dataset} is a subset of the DukeMTMC dataset \cite{ristani2016performance}, which is divided into 16,522 training images for 702 identities and 19,889 test images for 702 identities. Each image is annotated with 23 labels as we described. 
    
\textbf{The PETA dataset} is a large person attribute recognition dataset that annotated with 61 binary attributes and 4 multi-class attributes for 19,000 images. Following  \cite{deng2014pedestrian}, 35 most important and interesting attributes are used in our experiments. 
Since most identities have a few training images, and some only have one training image, PETA is not an ideal testbed for re-ID deep learning research. In this paper, to evaluate our method on PETA, we re-split the dataset for the re-ID task. We use 17,100 images of 4,981 identities for the experiment. In our new split, 9,500 images of 4,558 identities are used for training, 423 images are used for the query, and 7,177 images are used for the gallery.

\textbf{Evaluation metrics.} For the person re-ID task, the Cumulative Matching Characteristic (CMC) curve and the mean average precision (mAP) are used for evaluation. In the experiments, we use the evaluation package publicly available in \cite{Zheng_2015_ICCV,Zheng_2017_ICCV}. 
For the attribute recognition task, we test the classification accuracy for each attribute. The gallery images are used as the testing set. When testing the attribute prediction on Market-1501, we omit the distractor (background) and junks images, since they do not have attribute labels. We report the averaged accuracy of all these attribute predictions as the overall attribute prediction accuracy. 
    
\subsection{Implementation details}
In the experiments, we adopt ResNet-50 \cite{he2016deep} and CaffeNet \cite{jia2014caffe} as the CNN backbone, respectively. The network is initialized by ImageNet \cite{russakovsky2015imagenet} pre-trained models. Taking ResNet-50 for example, we append a 512-dim fully connected layer followed by Batch Normalization, a dropout layer with the drop rate of 0.5 and ReLU, after the “pool5” layer. The 512-dim fully connected layer is then concatenated with the 27-dim (for Market-1501) attribute prediction score.  The 539-dim (512+27) feature is used for identity classification. The experiment based on the CaffeNet is conducted similarly.
Finally, the classification layer with $k$ class nodes is used to predict the identity.
For each attribute, we adopt a fully connected layer after the "pool5" layer as the classifier for attribute prediction.
When evaluating the APR network for the re-ID task, we take the vertical concatenation of the embedded feature and the re-weighted attribute predictions as the final feature representation for each image.

Following \cite{zheng2017discriminatively}, we adopt a similar training strategy. Specifically, when using ResNet-50, we set the number of epochs to 60. The batch size is set to 32. Learning rate is initialized to 0.01 and changed to 0.001 in the last 20 epochs. 
For CaffeNet, the number of epochs is set to 110. For the first 100 epochs, the learning rate is 0.1 and changed to 0.01 in the last ten epochs. The batch size is set to 128. Randomly cropping and horizontal flipping are applied on the input images during training.

\subsection{Evaluation of Person Re-ID task} \label{sec:exp_reid}
  \setlength{\tabcolsep}{6pt}
  \begin{table*}[!t]
      \footnotesize
      \renewcommand{\arraystretch}{1.0}
      \begin{center}
            \caption{Comparison with state of the art on Market-1501. ``-'' indicates the papers use hand-crafted features, ``*'' indicates the papers use self-designed networks. ``w/o ARM'' denotes APR without the attribute re-weighting module. ``w/o attri'' denotes APR without the attribute recognition loss.}
      \begin{tabular}{l|c|c|cccc}
          \hline
          Methods&Publish&Backbone&Rank-1 & Rank-5 &Rank-10& mAP\\
          \hline
MBC \cite{ustinova2017multi}&AVSS2017&*&45.56&67&76&26.11\\
SML \cite{jose2016scalable}&ECCV2016&-&45.16&68.12&76&-\\
SL \cite{chen2016similarity}&CVPR2016&-&51.9&-&-&26.35\\
Attri \cite{matsukawa2016person}&ICPR2016&AlexNet&58.84&-&-&33.04\\
S-CNN \cite{varior2016gated}&ECCV2016&*&65.88&-&-&39.55\\
2Stream \cite{zheng2017discriminatively}&TOMM2017&Res50 &79.51&90.91&94.09&59.87\\
Cont-aware \cite{Li_2017_CVPR}&CVPR2017&*&80.31&-&-&57.53\\
Part-align \cite{Zhao_2017_ICCV}&ICCV2017&GoogLeNet&81.0&92.0&94.7&63.4\\ 
SVDNet \cite{Sun_2017_ICCV}&ICCV2017&Res50&82.3&92.3 &95.2&62.1\\
GAN \cite{Zheng_2017_ICCV}&ICCV2017&Res50&83.97&-&-&66.07\\
\re{EBB\cite{Tian_2018_CVPR}}&\re{CVPR2018}&Inception&\re{81.2} &\re{94.6}&\re{97.0}& \re{-}\\ 
\re{DSR\cite{He_2018_CVPR}}&\re{CVPR2018}&Res50&\re{82.72} &\re{-}&\re{-}& \re{61.25}\\ 
\re{AACN\cite{xu2018attention}}&\re{CVPR2018}&GoogLeNet&\re{85.90} &\re{-}&\re{-}& \re{66.87}\\ 
                \hline
                Baseline~1 &-&CaffeNet&54.76&73.28&82.04&28.75\\
                Baseline~1 &-&Res50&80.16&92.03&94.98&57.82 \\
                Baseline 2 &-&Res50&49.76&70.07&77.767&23.95\\
                \hline
                APR&-&CaffeNet&59.32&78.26&85.03&32.85\\             
                \re{APR (w/o attri)}&\re{-}&Res50&\re{81.03}&\re{91.29}&\re{94.28} &\re{58.74} \\
                APR (w/o ARM)&-&Res50&85.71&94.32&96.46 &66.59 \\
                APR &-&Res50&\textbf{87.04} &\textbf{95.10} &\textbf{96.42} &\textbf{66.89}\\
                \hline
            \end{tabular}
            \label{table:market}
        \end{center}
    \end{table*}
    
    \setlength{\tabcolsep}{10pt}
    \begin{table}
    \small
    
      \footnotesize
      \renewcommand{\arraystretch}{1.0}
        \begin{center}
        \caption{Comparison with the state of the art on DukeMTMC-reID with ResNet-50. - the respective papers use hand-crafted feature. Rank-1 accuracy (\%) and mAP (\%) are shown. ``w/o ARM'' denotes APR without the Attribute Re-weighting Module.}
            \begin{tabular}{l|c|cc}
                \hline
                Methods & Backbone&Rank-1 & mAP \\
                \hline
                BoW+kissme \cite{Zheng_2015_ICCV} & -&25.13 & 12.17 \\
                LOMO+XQDA \cite{liao2015person} & -&30.75 & 17.04\\
                AttrCombine \cite{matsukawa2016person}&AlexNet&53.87&33.35\\
                GAN  \cite{Zheng_2017_ICCV} &Res50& 67.68  & 47.13\\
                SVDNet \cite{Sun_2017_ICCV}    &Res50&\textbf{76.7}&\textbf{56.8}\\
         
                \hline
                Baseline~1  &Res50& 64.22 & 43.50\\ 
                Baseline 2  &Res50& 46.14 & 24.17\\ 
                \hline
                APR (w/o ARM)  &Res50&73.56 & 54.79 \\
                APR &Res50&73.92 & 55.56 \\
                \hline
            \end{tabular}
        \label{table:DukeMTMC-reID}
        \end{center}
    \end{table}
    
    \setlength{\tabcolsep}{5pt}
    \begin{table}[!t]
        \small
      \footnotesize
        \renewcommand{\arraystretch}{1.0}
            \caption{Person reID performance on PETA with ResNet-50. Rank-1 accuracy (\%) and mAP (\%) are shown. ``w/o ARM'' denotes APR without the Attribute Re-weighting Module.}
        \begin{center}
            \begin{tabular}{l|c|cccc}
            \hline
            Methods&Backbone&Rank-1 & Rank-5 &Rank-10& mAP\\
            \hline
            Baseline~1   &Res50& 53.90 & 68.32 & 73.04 &  49.60\\ 
            Baseline 2   &Res50& 43.30 & 60.08 & 70.23 & 39.52\\ 
            \hline
            APR (w/o ARM)   &Res50& 56.91& 72.10 &78.48 &  53.81 \\
            APR   &Res50& \textbf{58.05} & \textbf{78.34} & \textbf{75.73} &  \textbf{55.84} \\
            \hline
            \end{tabular}
            \label{table:peta1}
        \end{center}
    \end{table}
\subsubsection{Comparison with the state-of-the-art methods} The comparison with the state-of-the-art algorithms on Market-1501 and DukeMTMC-reID is shown in Table \ref{table:market} and Table \ref{table:DukeMTMC-reID}, respectively. On Market-1501, we obtain \textbf{rank-1 = 87.04\%, mAP = 66.89\%} by APR using the ResNet-50 model. We achieve the best rank-1 accuracy and mAP among the competing methods.  On DukeMTMC-reID, our results are \textbf{rank-1 = 73.92\% and mAP = 55.56\%} by APR using ResNet-50. Our method is thus shown to compare favorably with the state-of-the-art methods. 

\subsubsection{Comparison with the baselines}
Results on the three datasets are shown in Table \ref{table:market} Table \ref{table:DukeMTMC-reID} and Table \ref{table:peta1}. 
    
First, we observe that Baseline 2 (ARN) yields decent re-ID performance, \emph{e.g.}, a rank-1 accuracy of 49.76\% using ResNet-50 on Market-1501. 
Note that Baseline~2 only utilizes attribute annotations without ID labels. This illustrates that attributes are capable of discriminating between different persons. 

Second, by integrating the advantages in Baseline~1 and Baseline 2, our method exceed the two baselines by a large margin. For example,  when using ResNet-50, the rank-1 improvement on Market-1501 over Baseline~1 and Baseline 2 is 6.88\% and 37.28\%, respectively. On DukeMTMC-reID, APR achieves 9.7\% and 27.78\% improvement over Baseline~1 and Baseline 2 in rank-1 accuracy. The consistent finding also holds for PETA, \emph{i.e.}, we observe improvements of 4.15\% and 14.65\% over Baseline~1 and Baseline~2 in rank-1 accuracy, respectively. This demonstrates the complementary nature of the two baselines, \emph{i.e.}, identity and attribute learning. 
\re{We also observe that in Table \ref{table:market}, the performance of APR without attribute loss is slightly higher than that of B1. We believe that the slight improvement is lying on the difference of the network structure, that a Batch Normalization, a dropout layer and ReLU are further adopted in APR(w/o attri). However, the performance of both B1 and APR(w/o attri) still has a large margin between the performance of APR.  
}
    
Third, for both backbone models (\emph{i.e.,} CaffeNet and ResNet-50), APR yields consistent improvement. On Market-1501, we obtain 4.56\% and 6.88\% improvements in rank-1 accuracy over Baseline~1 with CaffeNet and ResNet-50, respectively.

\subsubsection{Ablation Studies}
\begin{figure}[!t]
        \begin{center}
            \includegraphics[width=\linewidth]{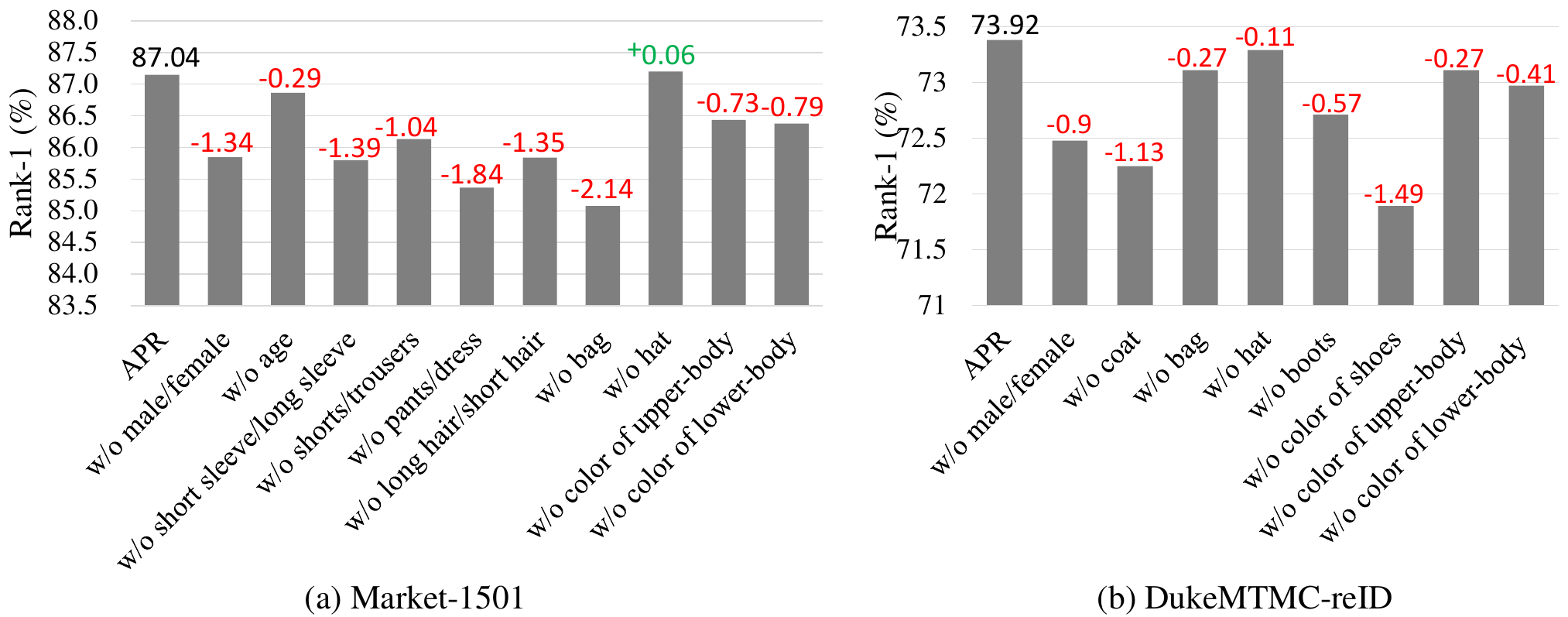}
        \end{center}
        \caption{Re-ID rank-1 accuracy on Market-1501 and DukeMTMC-reID. We remove one attribute from the system at a time. All the colors of upper-body clothing are viewed as one attribute here; the same goes for colors of lower-body clothing. Accuracy changes are indicated above the bars. }
        \label{fig:ablation}
\end{figure}
\textbf{Ablation study of attributes.} We evaluate the contribution of individual attributes on the re-ID performance. We remove each attribute from the APR system at one time, and the results on the two datasets are summarized in Fig. \ref{fig:ablation}. We find that most of the attributes on Market-1501 and DukeMTMC-reID are indispensable. The most influencing attribute on the two datasets are \emph{bag types} and \emph{the color of shoes}, which lead to a rank-1 decrease of 2.14\% and 1.49\% on the two datasets, respectively. This indicates that pedestrians of the two datasets have different appearances. The attribute of ``wearing a hat or not'' seems to exert a negative impact on the overall re-ID accuracy, but the impact is very small.
     
\textbf{The effectiveness of the Attribute Re-weighting Module.} We test APR with and without Attribute Re-weighting Module on the three re-ID datasets, and the results are shown in Table \ref{table:market}, Table \ref{table:DukeMTMC-reID} and Table \ref{table:peta1}. We observe performance improvement by using the Attribute Re-weighting Module for all the datasets. For Market-1501 with ResNet-50 as the backbone, the rank-1 and mAP improvements are 1.33\% and 0.30\%, respectively. For DukeMTMC-reID, the improvements are 0.36\% and 0.74\%, respectively. For PETA, we observe improvements of 1.14\% and 2.03\% in rank-1 and mAP, respectively. The improvement is consistent on all experiments.

\subsubsection{Algorithm Analysis}
    \begin{figure}[t]
        \begin{center}
            \includegraphics[width=0.8\linewidth]{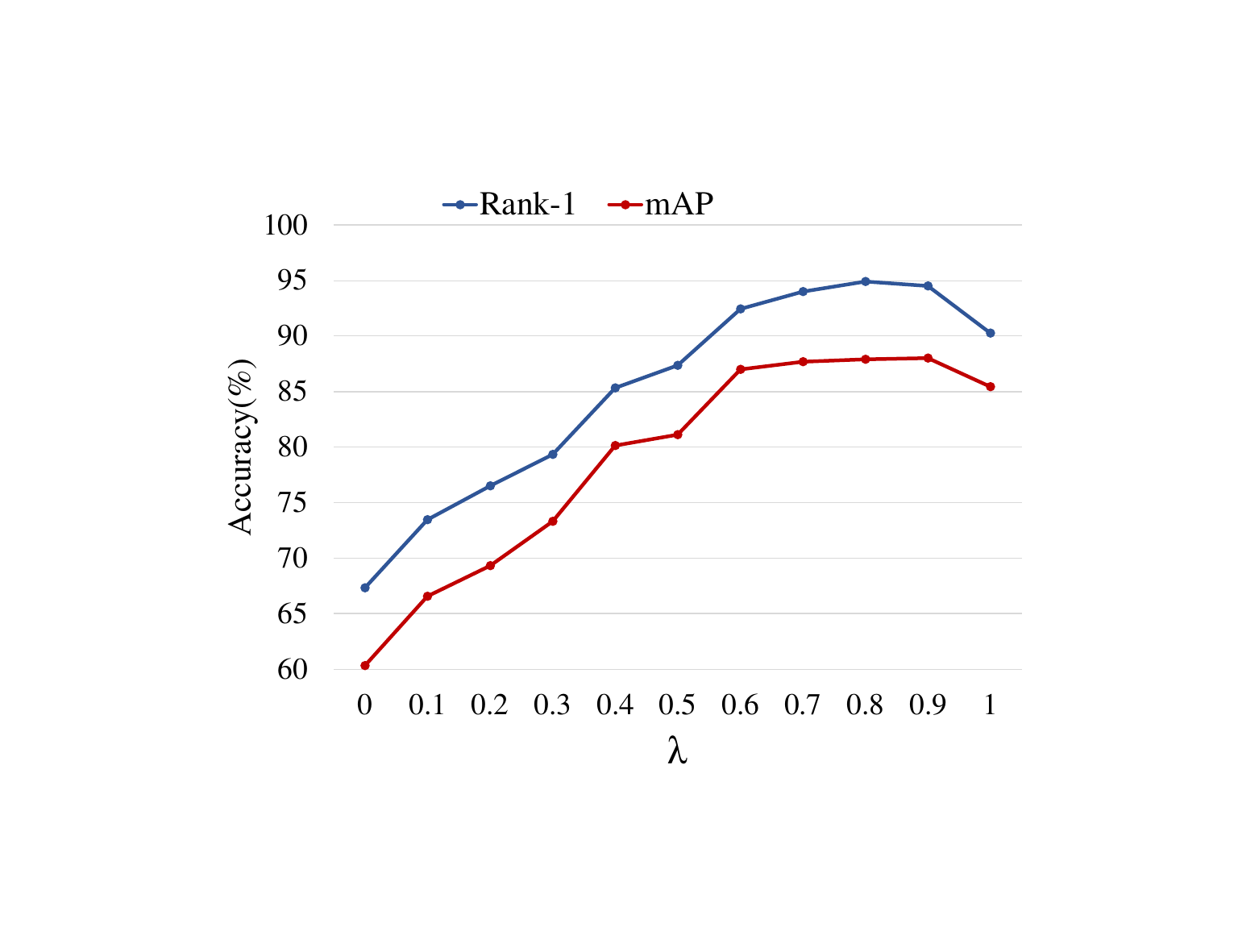}
        \end{center}
        \caption{ The re-ID performance (rank-1 accuracy and mAP) curves on the validation set of Market-1501 with different values of parameter $\lambda$ in Eq. \eqref{eq:APR_objective_function}. According to the performance curves, we set $\lambda = 0.9$ for all the experiment on Market-1501, DukeMTMC-reID and PETA.
        }
        \label{fig:parameter}
    \end{figure}
\textbf{Parameter validation.}

We validate the parameter ${\lambda}$ of APR on the validation set of Market-1501. ${\lambda}$ is a key parameter balancing the contribution of the identification loss and attribute recognition loss (Eq. \ref{eq:APR_objective_function}). When ${\lambda}$ becomes larger, person identity classification will play a more important role. Re-ID accuracy on the validation set of Market-1501 with different values of the parameter ${\lambda}$ is presented in Fig. \ref{fig:parameter}. We observe that when ${\lambda}$ changes from 0 to 0.9, the rank-1 accuracy and mAP gradually increase from 67.33\% and 60.32\% to 94.52\% and 88.03\%, respectively. It indicates the importance of identity label in the re-ID task. When ${\lambda}$ increases to 1, the rank-1 accuracy and mAP of the model decrease to 90.25\% and 85.44\% respectively, which indicates the effectiveness of attributes. The best re-ID performance is obtained when ${\lambda} = 0.9$. Therefore, we use  ${\lambda}= 0.9$ for APR in all the following experiments

\begin{figure} \centering 
        \subfigure { \label{fig:a} 
            \includegraphics[width=0.46\columnwidth]{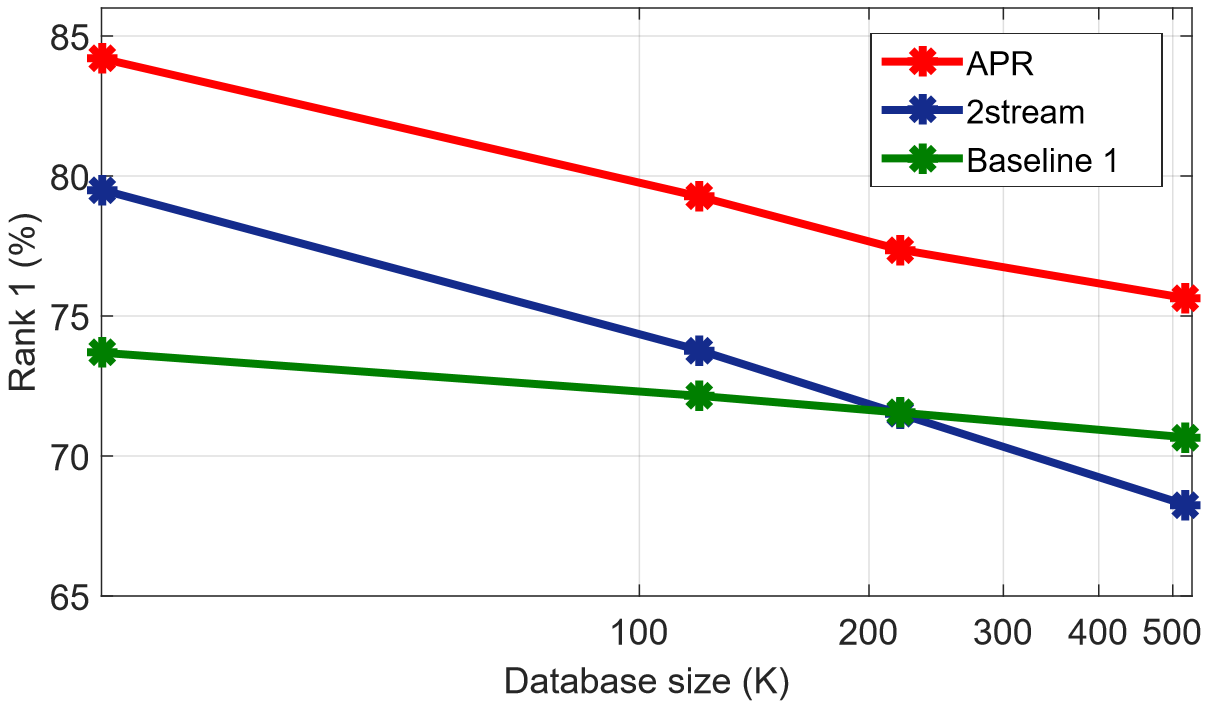} 
        } 
        \subfigure { \label{fig:b} 
            \includegraphics[width=0.46\columnwidth]{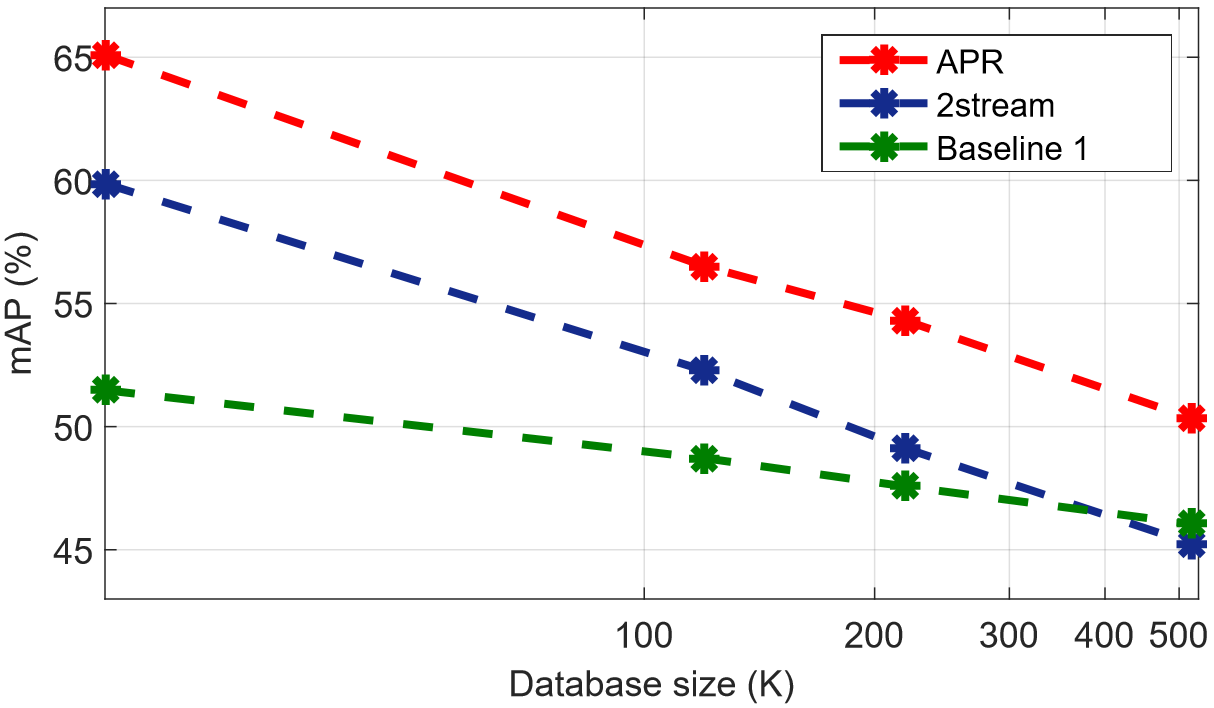} 
        } 
        \caption{Re-ID accuracy on the Market-1501+500k dataset. (Left:) rank-1 accuracy. (Right:) mean average precision. We compare our method with 2stream \cite{zheng2017discriminatively} and Baseline~1. As the number of images in the database increases, the accuracy of the three methods declines. However, APR remains the best performance.} 
        \label{fig:distractor} 
    \end{figure}

 \textbf{Robustness of the learned representation in the Wild.} To validate whether the proposed method still works under practical conditions, we report results on the Market-1501+500k dataset. The 500k distractor dataset is composed of background images and a large number of irrelevant pedestrians. The re-ID accuracy of our APR model with ResNet-50 on this dataset is presented in Fig. \ref{fig:distractor}. It can be expected that the re-ID accuracy drops as the gallery gets larger due to more distractors. The results further show that our method outperforms both \cite{zheng2017discriminatively} and Baseline~1. \re{However, the rank-1 accuracy of the proposed method drops faster than that of Baseline1. We think that the Baseline 1 may be able to retrieve the ground truths of easy queries, but APR could retrieve the ground truths of both the easy and hard queries. When increasing the number of images in the gallery, the easy query images can still be handled by both of the baseline and APR. However, the hard query sample can be harder to retrieve. Thus, the performance of APR drops faster.}

\subsubsection{Accelerating the retrieval process.}  

\begin{figure}[!t]
        \begin{center}
            \includegraphics[width=3.5in]{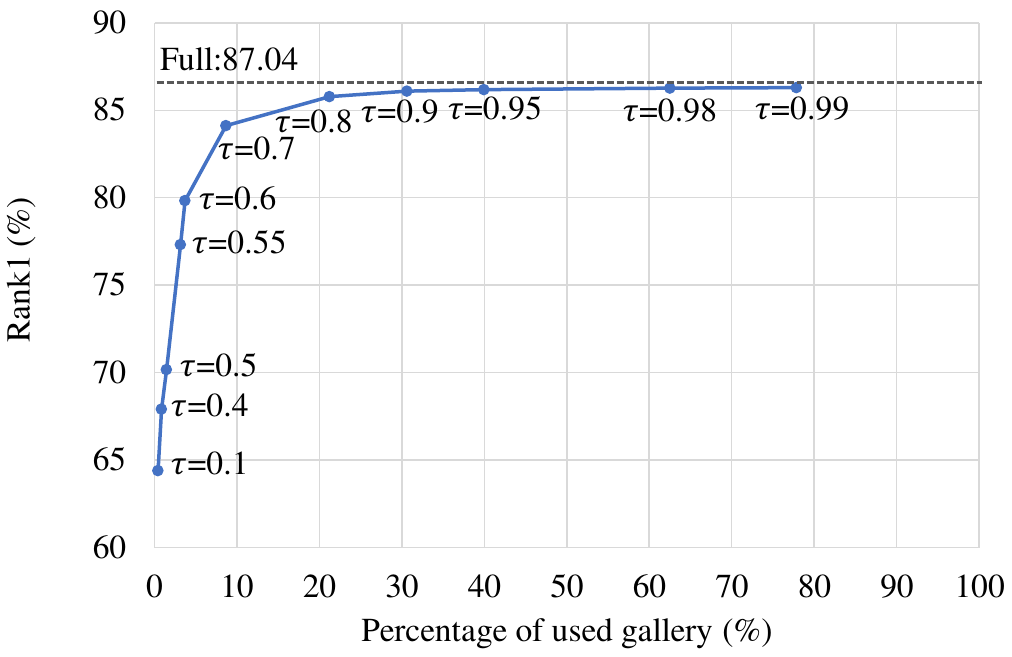}
        \end{center}
        \caption{Re-ID rank-1 accuracy curve on Market-1501 when using attributes to accelerate the retrieval process. For a query, we only take the gallery data with the same reliable attributes into consideration. The X-axis stands for different percentages of the \emph{remaining} gallery data when using different filtering threshold values.  Note that APR could speed up the retrieval process by nearly ten times (only 8.68\% gallery data remains) with only a slight accuracy drop of 2.92\%.}
        \label{fig:attri_filter}
\end{figure}

Fig. \ref{fig:attri_filter} illustrates the re-ID performance under different percentages of remaining gallery data. The number of remaining gallery images is controlled by the threshold $\tau$. It helps indicate if an attribute is reliable. As $\tau$ increases, attributes with higher confidence score are taken as reliable ones to wipe out gallery images, and the number of remaining gallery images increases.
As the percentage of remaining gallery data decreases from 78\% to 8.7\%, the rank-1 accuracy for re-ID decrease very slowly. When we try a more aggressive speedup, the performance drops quickly. For example, we observe an accuracy drop of 21.79\% when we use only 0.5\% gallery images. Note that with the remaining 8.68\% gallery data, we still achieve 84.12\% on rank-1 accuracy, which is close to the original result 87.04\%.

In practice, most of the time is spent on calculating the distances between the query feature and the features of remained gallery images. In Market-1501, for instance, there are 3,368 queries and 19,732 gallery images. Without the acceleration, the testing process takes 919.86s (0.273s per query), using an Intel i7-6850K CPU. With acceleration, there are only about 2,000 images remaining in the gallery for each query, and it costs only 90.26s (0.026s per query) for testing. Although the saved time may be slight in the academic dataset, in real-world applications which involves a large amount of data, efficiency could be an important advantage.

\subsection{Evaluation of Attribute Recognition}\label{sec:eva_attribute}

We test attribute recognition on the galleries of the Market-1501, DukeMTMC-reID, PETA in Table \ref{table:compare}, Table \ref{table:compare_att_DukeMTMC-reID}, and Table \ref{table:peta2}, respectively. 
We also evaluate our method on the CUB\_200\_2011 dataset, which contains 11,788 images of 200 bird classes. Each category is annotated with 312 attributes, which are divided into 28 groups and are used as 28 multi-class attributes in our experiments. The result is shown in Table \ref{table:cub2}. By comparing the results of APR and Baseline 2 (ARN), two conclusions can be drawn:
    
    \setlength{\tabcolsep}{2.75pt}
    \begin{table*}[!t]
    	\small
            \caption{Attribute recognition accuracy on Market-1501. In ``APR'', parameter $\lambda$ is optimized in Fig. \ref{fig:parameter}. ``L.slv'', ``L.low'', ``S.clth'', ``B.pack'', ``H.bag'', ``C.up'', ``C.low'' denote \emph{length of sleeve}, \emph{length of lower-body clothing}, \emph{style of clothing}, \emph{backpack}, \emph{handbag}, \emph{color of upper-body clothing} and \emph{color of lower-body clothing}, \emph{resp}. ``B2'' denotes Baseline~2 (ARN).}
        \begin{center}
            \begin{tabular}{l|cccccccccccc|c}
                \hline
                &gender & age&hair &L.slv& L.low &S.clth&B.pack&H.bag&bag&hat&C.up&C.low& Avg\\
                \hline
                B2 &87.5&85.8&84.2&93.5&93.6&93.6&86.6&88.1&78.6&97.0&72.4 &71.7 &86.0\\
                APR&88.9&88.6&84.4&93.6&93.7&92.8&84.9&90.4&76.4&97.1& 74.0&73.8 &86.6\\
                \hline
            \end{tabular}
            \label{table:compare}
        \end{center}
    \end{table*}
    
    \setlength{\tabcolsep}{3.5pt}
    \begin{table*}[!t]
        \small
        \caption{Attribute recognition accuracy on DukeMTMC-reID. ``L.up'', ``B.pack'', ``H.bag'', ``C.shoes'', ``C.up'', ``C.low'' denote \emph{length of sleeve}, \emph{backpack}, \emph{handbag}, \emph{color of shoes}, \emph{color of upper-body clothing} and \emph{color of lower-body clothing}, \emph{resp}. ``B2'' denotes Baseline~2 (ARN).}
        \renewcommand{\arraystretch}{1.0}
        \begin{center}
            \begin{tabular}{l|cccccccccc|c}
                \hline
                &gender & hat&boots &L.up &B.pack&H.bag&bag&C.shoes&C.up&C.low& Avg\\
                \hline
                B2 &82.0& 85.5 &88.3&86.2&77.5&92.3&82.2&87.6&73.4&68.3 &82.3\\
                APR &84.2&87.6&87.5&88.4&75.8&93.4&82.9& 89.7& 74.2&69.9&83.4\\
                \hline
            \end{tabular}
            \label{table:compare_att_DukeMTMC-reID}
        \end{center}
    \end{table*}

\setlength{\tabcolsep}{14pt}
\begin{table}[!htb]
   \caption{Attribute recognition accuracy on PETA.}
   \vspace{-4mm}
    \begin{center}
        \small
        \begin{tabular}{l|c}
            \hline
            Attribute & mean accuracy\\
            \hline
            MRFr2 \cite{deng2014pedestrian}&71.1\\
            ACN \cite{sudowe2015person}&81.15\\
            MVA \cite{schumann2017person}&84.61\\
            \hline
            Baseline 2&84.45\\
            APR& \textbf{84.94}\\
            \hline
        \end{tabular}
    \end{center}
    \label{table:peta2}
\end{table}
    
    \setlength{\tabcolsep}{14pt}
\begin{table}[!htb]
    \caption{Attribute recognition accuracy on CUB\_200\_2011.}
    \vspace{-4mm}
    \begin{center}
        \small
        \begin{tabular}{l|c}
            \hline
            Methods & mean accuracy \\
            \hline
            Baseline 2 &87.31 \\
            APR&\textbf{89.12} \\
            \hline
        \end{tabular}
    \end{center}
    \label{table:cub2}
\end{table}

First, on all datasets, the overall attribute recognition accuracy is improved by the proposed APR network to some extent. The improvements are 0.26\%, 0.08\%, 0.2\% and 1.58\% on Market-1501, DukeMTMC-reID, PETA and CUB\_200\_2011, respectively. So overall speaking, the integration of identity classification introduces some degree of complementary information and helps in learning a more discriminative attribute model. Also, note that we achieve the best attribute recognition result on PETA among the state-of-the-art.
    
Second, we observe that the recognition rate of some attributes decreases for APR, such as \emph{hair} and \emph{B.pack} in Market-1501. However, Fig. \ref{fig:ablation} demonstrates that these attributes are necessary for improving re-ID performance. The reason probably lies in the multi-task nature of APR. Since the model is optimized for re-ID (Fig. \ref{fig:parameter}), ambiguous images of certain attributes may be incorrectly predicted. Nevertheless, the improvement on the two datasets is still encouraging and further investigations should be critical.

\section{Conclusions and Future Work}
In this paper, we mainly discuss how re-ID is improved by the integration of attribute learning.
Based on the complementary of attribute labels and ID labels, we propose an attribute-person recognition (APR) network, which learns a re-ID embedding and predicts the pedestrian attributes under the same framework.
We systematically investigate how the person re-ID and attribute recognition benefit each other. 
In addition, we re-weight the attribute predictions considering the dependencies and correlations among attributes of a person.
To show the effectiveness of our method, we have annotated attribute labels on two large-scale re-ID datasets. The experimental results on two large-scale re-ID benchmarks demonstrate that by learning a more discriminative representation, APR achieves competitive re-ID performance compared with the state-of-the-art methods. 
We additionally use APR to accelerate the retrieval process of re-ID more than three times with a minor accuracy drop of 1.26\% on Market-1501. For attribute recognition, we also observe an overall precision improvement using APR.

Pedestrian attributes provide a different view of the person re-identification problem. As a mid-level feature, attributes are more robust to environment changes, such as the background and illumination. In the future, we will first investigate the transferability and scalability of pedestrian attributes. For instance, we could adapt the attribute model learned on Market-1501 to other pedestrian datasets. Second, attributes provide a bridge to the image-text understanding. We will investigate a system using attributes to retrieve the relevant pedestrian images. It is useful in solving specific re-ID problems, in which the query image is missing and can be described by attributes.

\bibliography{mybibfile}

\end{document}